\documentclass[runningheads]{llncs}

\usepackage{eccv}

\usepackage{eccvabbrv}

\usepackage{graphicx}
\usepackage{booktabs}

\usepackage[accsupp]{axessibility}  

\usepackage{hyperref}
\usepackage{multicol}
\usepackage{multirow}
\usepackage{booktabs}
\usepackage{graphicx}

\usepackage{orcidlink}

\begin{document}


\title{MapRoute++: Surrogate-Guided Semantic Routing for Visual Concept Unlearning}





\titlerunning{MapRoute++}

\author{Ashok Urlana\inst{1, 2} \and
L. D. M. S. Sai Teja\inst{3} \and
 Vivek Hruday Kavuri\inst{1} \and 
 Ponnurangam Kumaraguru\inst{1}}

\authorrunning{Urlana et al.}


\institute{$^1$IIIT Hyderabad, $^2$TCS Research, India, $^3$NIT Silchar}

\maketitle

\begin{abstract}
We present our submission to Task 3 of the Gen$\mu$ 2.0 Challenge on visual concept unlearning. Building on MapRoute, we introduce task-specific training objectives, richer concept representations, and semantic routing for concept-specific mapper selection. Our approach improves robust concept removal while preserving unrelated and semantically adjacent concepts. On the official benchmark, evaluated using the Erasing-Retention-Robustness (ERR) metric on Stable Diffusion v1.4, our method outperforms the state-of-the-art baseline by 12.1\% on average across the five concept categories, achieving substantial gains. The code and dataset are available on \href{https://github.com/ashokurlana/MapRoute-plus-plus}{GitHub}.
\end{abstract}

\section{Introduction}
\label{sec:intro}

Recent text-to-image diffusion models \cite{chang2023muse, dhariwal2021diffusion, kim2022diffusionclip, ramesh2022hierarchical, rombach2022high, saharia2022photorealistic} have achieved remarkable performance in image generation for diverse text prompts, enabling applications such as art generation, content creation, and data augmentation. However, these models also learn undesirable concepts, including copyrighted content and societal biases, raising concerns about fairness and responsible deployment \cite{mishkin2022risks, rando2022red, schramowski2023safe, somepalli2023diffusion}. Visual concept unlearning, or concept erasure, seeks to selectively remove such concepts while preserving the model's ability to generate unrelated and semantically adjacent concepts \cite{belrose2023leace, pham2024robust, radford2021learning}.

In this work, we address the visual concept unlearning task of the Gen$\mu$ 2.0 Challenge\footnote{https://iab-iitj.github.io/genmu2026/index.html}, which evaluates concept erasure on Stable Diffusion v1.4 \cite{rombach2022high} using the ERR (Erasing--Retention--Robustness) score under direct, indirect, and adversarial prompts. Existing concept erasure approaches can be broadly categorized into optimization-based methods, such as ESD \cite{gandikota2023erasingconceptsdiffusionmodels}, CRCE \cite{xue2025crcecoreferenceretentionconcepterasure}, CORE \cite{zhu2024choose}, AGE \cite{bui2025fantastictargetsconcepterasure}, and FADE \cite{thakral2025fine}, and modular approaches, including SPM \cite{lyu2024onedimensionaladapterruleall}, Receler \cite{huang2024receler}, and MapRoute \cite{Li_2026_CVPR}. Building on MapRoute, we enhance concept-specific mapper training with task-specific objectives, diverse textual concept representations, and semantic routing to achieve robust concept removal while preserving generation quality.

\section{Challenge Dataset}
As part of the Gen$\mu$ 2.0 Challenge on visual concept unlearning, the organizers released a challenge dataset\footnote{https://www.kaggle.com/competitions/gen-challenge-u-me-workshop-iccv-25/data}. The dataset comprises 20 concepts spanning five categories: three Object concepts (Barberton Daisy, Golf Ball, and Apple Fruit), two Animal concepts (Blue Jay and Labrador Retriever), five Style concepts (Van Gogh, Doodle, Neon, Monet, and Sketch), five Scene concepts (Wedding, Sunset, Rainfall, Aurora Borialis, and Scenerie), and five Action concepts (Sleeping, Walking, Eating, Dancing, and Jumping). Moreover, the dataset includes direct, indirect, and adversarial prompts, plus adjacent concepts (semantically similar to the target) and retained concepts, to assess the unlearned model's generalization.


\section{Methodology}
The MapRoute++ framework aims to erase the target concepts without compromising the model's utility. The framework consists of three components: 1) a lightweight per-concept mapper module inserted between the frozen text encoder and the U-Net denoiser in the diffusion model, 2) a two-stage training procedure to suppress a target concept while preserving the semantic representation of unrelated concepts, and 3) an input-conditioned routing mechanism that dynamically selects the appropriate trained module at inference time.
\begin{figure}[t]
    \centering
    \includegraphics[width=1.0\linewidth]{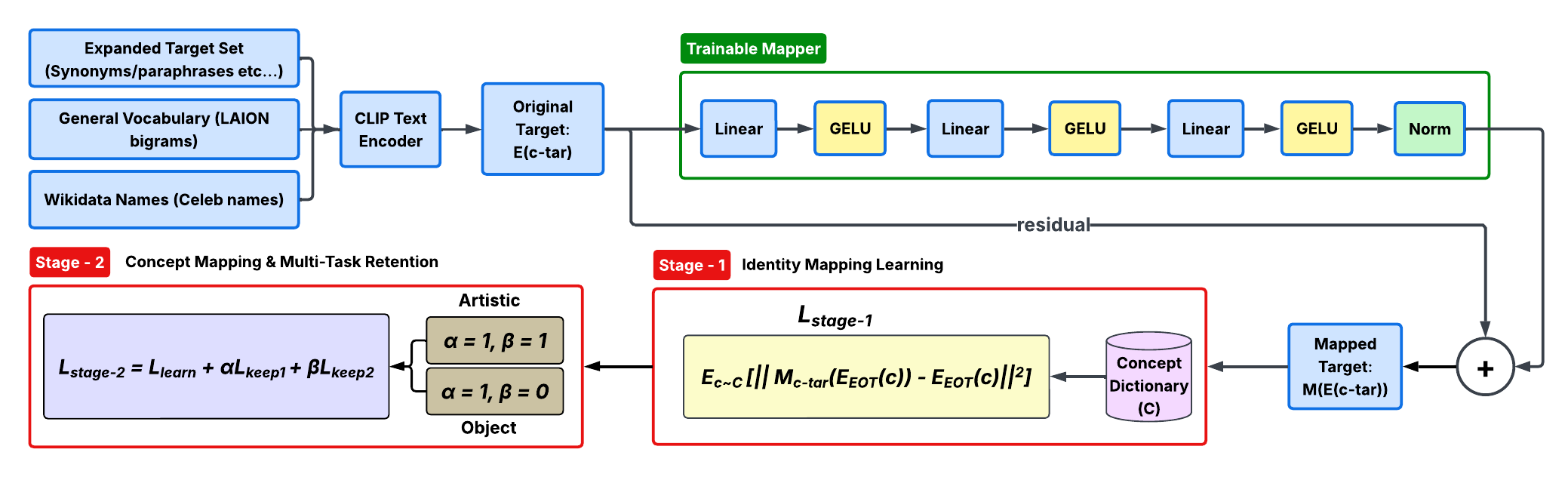}
    \caption{MapRoute++ uses a two-stage residual MLP to redirect concepts to surrogates in frozen diffusion models, modifying the sequence-level EOT token embedding for targeted suppression while preserving semantic fidelity.}
    \label{fig:pipeline}
\end{figure}

\subsection{Design of Concept-Specific Mapper Module}

Existing concept erasure techniques rely on concept-specific training or fine-tuning of diffusion models, both requiring high-quality paired data (e.g., ``target'' vs. ``surrogate'' concepts) to precisely suppress concepts, often degrading generation quality. To overcome these problems, as shown in Figure~\ref{fig:pipeline}, our approach, for each target concept, trains a lightweight residual multilayer perceptron (MLP) to transform the corresponding text embedding, while keeping embeddings of unrelated concepts largely unchanged.
Let $E(\cdot)$ denote the frozen text encoder and $c$ a text prompt. For each target concept $c_{tar}$, a dedicated module $M_{c_{tar}}$ is trained as a conditional identity mapping:
\begin{equation}
M_{c_{tar}}(E(c)) =
\begin{cases}
E(c_{sur}), & c = c_{tar} \\[-2pt]
E(c), & c \neq c_{tar}
\end{cases}
\tag{1}
\end{equation}
where $c_{sur}$ is the chosen substitute concept. The module leaves all embeddings unchanged except its assigned target, which it redirects toward the substitute. 
\subsection{Two-Stage Training Strategy}
The Mapper module is optimized in two stages, ensuring it suppresses only the intended concept while leaving all other semantics intact.

\noindent \textbf{Stage 1: Identity Pretraining.} $M_{c_{tar}}$ is first trained to reconstruct embeddings across a broad concept vocabulary $C$, establishing no changes in the embeddings as baseline before introducing any suppression behavior.
\begin{equation}
\mathcal{L}_{stage1} = \mathbb{E}_{c \in C}\Big[\, \| M_{c_{tar}}(E(c)) - E(c) \|_2^2 \,\Big]
\tag{2}
\end{equation}
In this stage, the mapper module behaves as an identity function for any input it has not been explicitly trained to alter.

\noindent \textbf{Stage 2: Target-Surrogate Mapping.} Building on this baseline, the module is trained to redirect the target concept's embedding toward that of a chosen surrogate concept: 
\begin{equation}
\mathcal{L}_{learn} =
\mathbb{E}_{c_{tar}}\Big[
\| M_{c_{tar}}(E(c_{tar})) - E(c_{sur}) \|_2^2
\Big]
\tag{3}
\end{equation}
To prevent this targeted update from disturbing unrelated concepts, two regularization terms are added. The $\mathcal{L}_{keep1}$ reinforces the identity mapping and $\mathcal{L}_{keep2}$ does the same over a curated set of proper names $W$, protecting celebrity and identity-related concepts from unintended erasure. Overall, Stage 2 objective is:


\begin{equation}
\mathcal{L}_{stage2} =
\mathcal{L}_{learn}
+ \alpha \cdot \mathcal{L}_{keep1}
+ \beta \cdot \mathcal{L}_{keep2}
\tag{4}
\end{equation}
where $\alpha, \beta$ (default $=1$) balance erasure strength against preservation of unrelated concepts. Together, the two stages allow the Mapper to achieve near-zero distortion on non-target concepts while reliably suppressing the designated one.

\noindent Following MapRoute~\cite{Li_2026_CVPR}, the surrogate concept $c_{sur}$ is not critical for erasure quality, since the mapper mainly learns to move the target embedding away from its original region rather than toward a specific destination. Unlike the original MapRoute, we adopt a task-specific objective for artistic styles, the full Stage 2 loss (Eq.~4) including $\mathcal{L}_{keep2}$ is retained to protect artist identities, for objects, $\mathcal{L}_{keep2}$ is omitted. Additionally, each concept is represented via multiple synonyms and paraphrases rather than a single keyword, enriching $\mathcal{L}_{learn}$ to encourage genuine semantic removal and better generalization to unseen or adversarial prompts.

\begin{table}[t]
\centering
\setlength{\tabcolsep}{7pt}
\caption{Comparison of Average ERR Score by Various Approaches. Best performing score per type is \textbf{bolded} and next best performing score is \underline{underlined}.}
\label{tab:err_by_category}

\begin{tabular}{lcccccc}
\toprule
\textbf{Type} & \textbf{ESD} & \textbf{CA} & \textbf{FMN} & \textbf{FADE} & \textbf{MapRoute} & \textbf{MapRoute++} \\
\midrule
Object & 0.470 & 0.471 & 0.487 & 0.598 & \textbf{0.919} & \underline{0.863} \\
Animal & 0.499 & 0.497 & 0.467 & 0.599 & \textbf{0.796}& \underline{0.604} \\
Style  & 0.478 & 0.509 & 0.482 & \textbf{0.609} & 0.494 & \underline{0.574} \\
Scene  & 0.461 & 0.422 & 0.443 & \underline{0.589} & 0.190 &\textbf{0.707} \\
Action & 0.477 & 0.460 & 0.403 & 0.607 & \textbf{0.871} &\underline{0.857} \\
\midrule
\textbf{Avg} & 0.477 & 0.472 & 0.456 & \underline{0.600} & 0.591&\textbf{0.721} \\
\bottomrule
\end{tabular}
\end{table}

\subsection{Dynamic Semantic Routing}
At inference, the framework computes the semantic similarity between the input prompt and all target concepts using the mean of the prompt's token embeddings, and sequentially applies the top-$k$ most relevant mapper modules. Although each mapper is trained on sentence-level text embeddings, it is applied token-wise to the full embedding sequence, enabling the learned semantic transformation to modify the cross-attention inputs while remaining fully compatible with the pretrained diffusion model.

\section{Experiments and Analysis}
\subsection{Experimental setup}
We conduct all our experiments on Stable Diffusion-v1.4 \cite{rombach2022high} model. We generate 20 images for each concept and obtain the ERR (Erasing-Retention-Robustness)\footnote{\url{https://shreyanshhub.github.io/GENMU-/evaluation.html}} score, an official metric taken from the Gen$\mu$ Challenge. For the LLaVA-based \cite{liu2024improved} ERR evaluation, every image is scored using the LLaVA\footnote{\url{https://huggingface.co/llava-hf/llava-v1.6-mistral-7b-hf}} model with a constrained yes/no question. We use the same question form for all axes:
\begin{quote}
\small
\texttt{Does this image show <concept description>? Answer with `yes' or `no'.}
\end{quote}
 For direct, indirect, and adversarial target prompts, success means that the target is no longer present, so the expected answer is \texttt{no}. For adjacent and retained prompts, success means that the non-target concept is preserved, so the expected answer is \texttt{yes}, as given in Table~\ref{tab:llava_prompt}. For example, the Blue Jay target is queried as ``a Blue Jay bird'', and Van Gogh is queried as ``Van Gogh painting style''.


\noindent Image-level binary decisions are aggregated across all concepts to compute five evaluation axes: target forgetting ($A_{\text{fgt}}$), retention of unrelated concepts ($A_{\text{ret}}$), retention of semantically adjacent concepts ($A_{\text{adj}}$), robustness to indirect prompts ($A_{\text{ind}}$), and robustness to adversarial prompts ($A_{\text{adv}}$). The final ERR score is computed as the harmonic mean of these five quantities, penalizing methods that achieve high performance on only a subset of the evaluation criteria. We conducted all experiments using two Nvidia GeForce RTX A6000 (48GB) GPUs and followed the same parameters as mentioned in the official MapRoute GitHub repo\footnote{\url{https://github.com/GG-li/MapRoute}} for all the experiments.

\begin{table}[t]
\centering
\setlength{\tabcolsep}{3.5pt}
\caption{Per-Concept Breakdown of MapRoute++ Evaluation Results. Note: Animals are combined in the objects category.}
\label{tab:per_concept_breakdown}
\begin{tabular}{cclcccccc}
\toprule
\# & Type & Concept
& $A_{fgt}$ & $A_{ret}$ & $A_{adj}$ & $A_{ind}$ & $A_{adv}$ & ERR \\
\midrule
1 & \multirow{5}{*}{\rotatebox[origin=c]{90}{Objects}}
  & Labrador Retriever & 1.000 & 0.9368 & 0.700 & 1.000 & 1.000 & 0.9098 \\
2 & & Barberton Daisy      & 0.900 & 0.9368 & 0.350 & 1.000 & 1.000 & 0.7107 \\
3 & & Blue Jay             & 1.000 & 0.9474 & 0.083 & 1.000 & 0.600 & 0.2990 \\
4 & & Golf Ball            & 0.950 & 0.9895 & 0.933 & 1.000 & 1.000 & 0.9738 \\
5 & & Apple Fruit          & 1.000 & 0.9342 & 0.683 & 1.000 & 1.000 & 0.9035 \\
\midrule
6 & \multirow{5}{*}{\rotatebox[origin=c]{90}{Artistic Styles}}
  & Van Gogh & 0.300 & 0.9368 & 1.000 & 0.250 & 0.300 & 0.3926 \\
7 & & Doodle  & 0.100 & 0.9500 & 1.000 & 1.000 & 1.000 & 0.3558 \\
8 & & Neon    & 0.850 & 0.9395 & 0.883 & 0.800 & 0.900 & 0.8720 \\
9 & & Monet   & 0.250 & 0.9368 & 1.000 & 0.450 & 0.550 & 0.4947 \\
10 & & Sketch & 0.700 & 0.9447 & 0.867 & 0.650 & 0.700 & 0.7567 \\
\midrule
11 & \multirow{5}{*}{\rotatebox[origin=c]{90}{Open Set}}
   & Wedding         & 1.000 & 0.9474 & 0.667 & 1.000 & 1.000 & 0.9000 \\
12 & & Sunset          & 0.200 & 0.9474 & 0.950 & 0.400 & 1.000 & 0.4713 \\
13 & & Rainfall        & 1.000 & 0.9421 & 0.333 & 0.950 & 1.000 & 0.7028 \\
14 & & Aurora Borealis & 1.000 & 0.9447 & 0.667 & 1.000 & 1.000 & 0.8995 \\
15 & & Scenerie        & 1.000 & 0.9395 & 0.983 & 0.300 & 0.400 & 0.5609 \\
\midrule
16 & \multirow{5}{*}{\rotatebox[origin=c]{90}{Activities}}
   & Sleeping & 1.000 & 0.9500 & 0.667 & 1.000 & 1.000 & 0.9005 \\
17 & & Walking  & 0.500 & 0.9553 & 0.883 & 0.400 & 1.000 & 0.6511 \\
18 & & Eating   & 0.950 & 0.9579 & 0.950 & 1.000 & 1.000 & 0.9710 \\
19 & & Dancing  & 0.750 & 0.9395 & 0.983 & 1.000 & 1.000 & 0.9234 \\
20 & & Jumping  & 0.700 & 0.9474 & 0.683 & 1.000 & 1.000 & 0.8407 \\
\midrule
\multicolumn{3}{c}{\textbf{Average}}
& \textbf{0.7575}
& \textbf{0.9462}
& \textbf{0.7633}
& \textbf{0.8100}
& \textbf{0.8725}
& \textbf{0.7245} \\
\bottomrule
\end{tabular}
\end{table}


\begin{table}[h]
\centering
\setlength{\tabcolsep}{5pt}
\caption{LLaVA prompt targets and success criteria used for ERR.}
\label{tab:llava_prompt}
\small
\begin{tabular}{lll}
\toprule
Prompt Type & Queried concept & Success answer \\
\midrule
Direct ($A_{\mathrm{fgt}})$ & Target concept & \texttt{no} \\
Indirect ($A_{\mathrm{ind}})$ & Target concept & \texttt{no} \\
Adversarial ($A_{\mathrm{adv}})$ & Target concept & \texttt{no} \\
Adjacent ($A_{\mathrm{adj}})$ & Adjacent concept & \texttt{yes} \\
Retention ($A_{\mathrm{ret}})$ & Retained concept & \texttt{yes} \\
\bottomrule
\end{tabular}
\end{table}

\subsection{Baselines Performance Analysis}
\label{sec:baseline_analysis}

We compare MapRoute++ against five recent concept erasure baselines provided in the Gen$\mu$ 2.0 Challenge: ESD \cite{gandikota2023erasing}, CA \cite{kumari2023ablating}, FMN \cite{zhang2024forget}, FADE \cite{thakral2025fine} and MapRoute \cite{Li_2026_CVPR}. Table~\ref{tab:err_by_category} presents the average ERR scores across the five distinct conceptual categories (Object, Animal, Style, Scene, and Action) along with the overall average performance. 

\noindent \textbf{Overall Superiority.} 
MapRoute++ achieves a new state-of-the-art overall average ERR score of 0.721, substantially outperforming the strongest baseline, FADE. This translates to a $12.1\%$ absolute improvement on average across all categories. In contrast, earlier methods such as ESD, CA, FMN and MapRoute struggle to balance the competing objectives of target forgetting, semantic preservation, and prompt robustness, clustering below the 0.600 mark. 

\noindent \textbf{Category-Specific Strengths.} 
The advantages of our surrogate-guided semantic routing are most pronounced in the \textbf{Object} and \textbf{Scene} categories. For scenes, MapRoute++ achieves an ERR of 0.707, vastly surpassing MapRoute by a margin of 50\%+ improvement. These results indicate that modifying the sequence-level EOT token embedding via our mapper is highly effective for localized, discrete concepts (e.g., \textit{Golf Ball}, \textit{Eating}) that do not dictate global image statistics. \\
\noindent \textbf{Limitations in Style Erasure.} 
The \textbf{Style} category remains the most challenging for our approach. As evidenced by our per-concept breakdown (Table~\ref{tab:per_concept_breakdown}) and qualitative failure cases (Section 5.1), artistic styles like \textit{Van Gogh} and \textit{Doodle} heavily rely on global texture, brushwork, and structural composition. Because MapRoute++ operates primarily on semantic token embeddings rather than fine-tuning the cross-attention weights of the U-Net denoiser, completely untangling pervasive stylistic features without aggressively degrading adjacent concepts proves difficult. Nonetheless, MapRoute++ still outperforms ESD, CA, FMN and MapRoute by comfortable margins even in this category.

\begin{figure}[t]
\centering
\setlength{\tabcolsep}{2pt}
\begin{tabular}{cccc}
\includegraphics[width=0.235\linewidth]{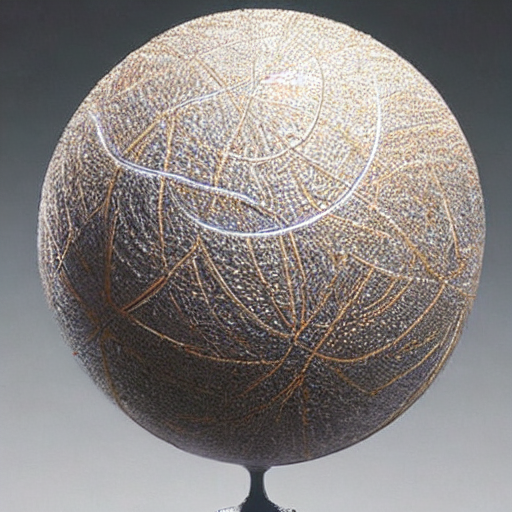} &
\includegraphics[width=0.235\linewidth]{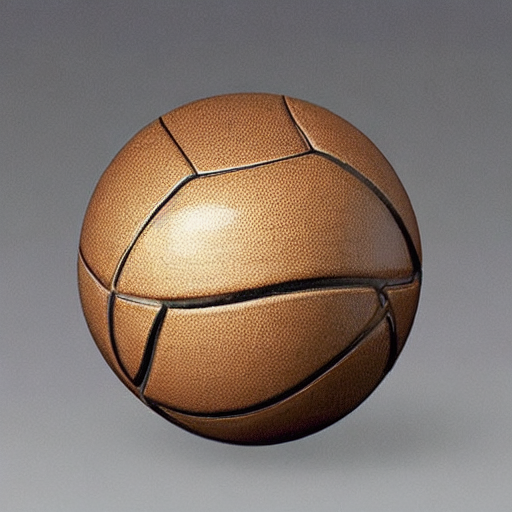} &
\includegraphics[width=0.235\linewidth]{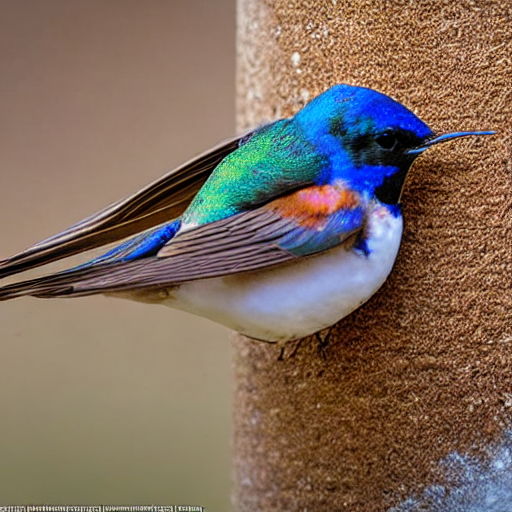} &
\includegraphics[width=0.235\linewidth]{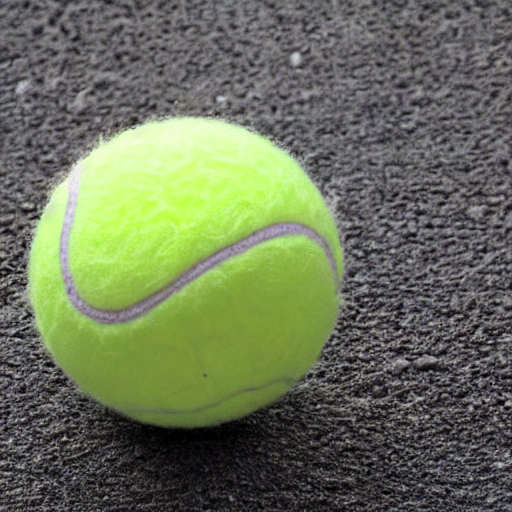} \\
{\scriptsize Golf Ball base} &
{\scriptsize Direct} &
{\scriptsize Adversarial} &
{\scriptsize Tennis Ball} \\

\includegraphics[width=0.235\linewidth]{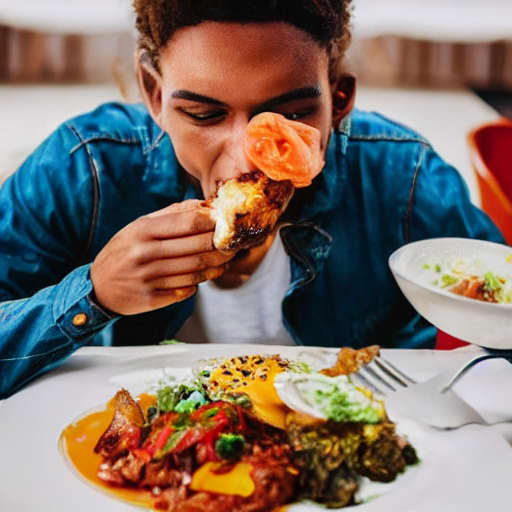} &
\includegraphics[width=0.235\linewidth]{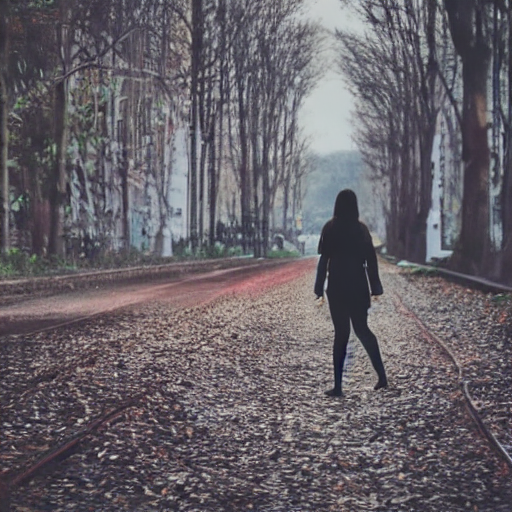} &
\includegraphics[width=0.235\linewidth]{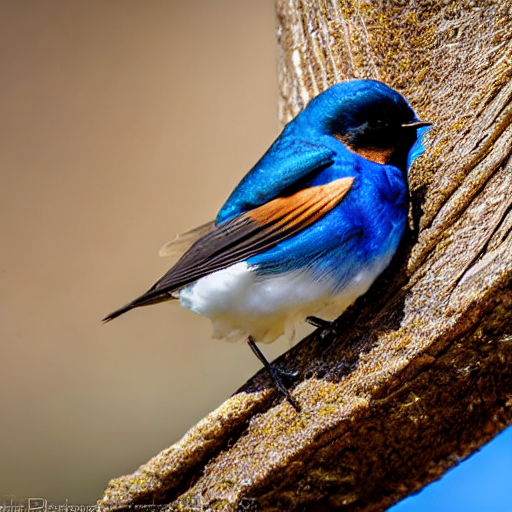} &
\includegraphics[width=0.235\linewidth]{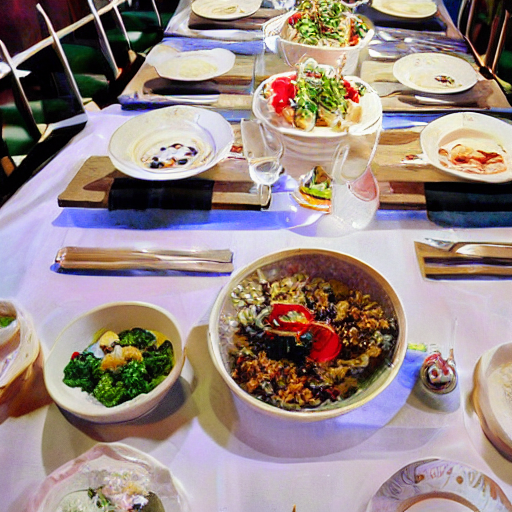} \\
{\scriptsize Eating base} &
{\scriptsize Direct} &
{\scriptsize Adversarial} &
{\scriptsize Dining} \\

\includegraphics[width=0.235\linewidth]{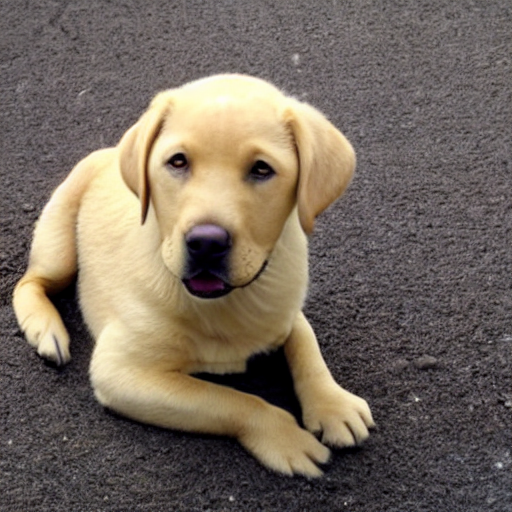} &
\includegraphics[width=0.235\linewidth]{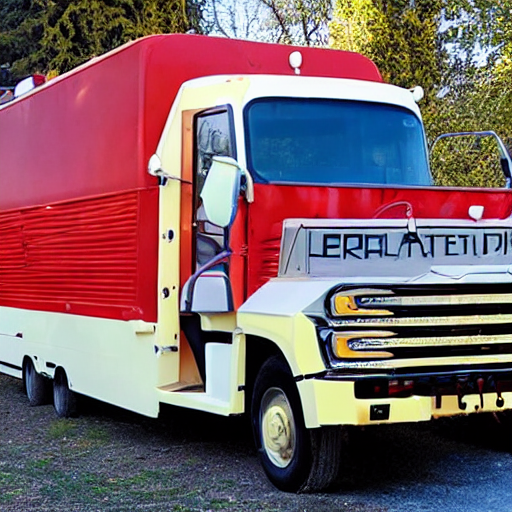} &
\includegraphics[width=0.235\linewidth]{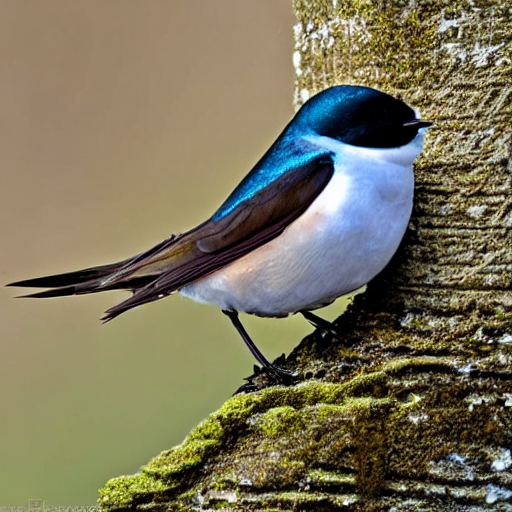} &
\includegraphics[width=0.235\linewidth]{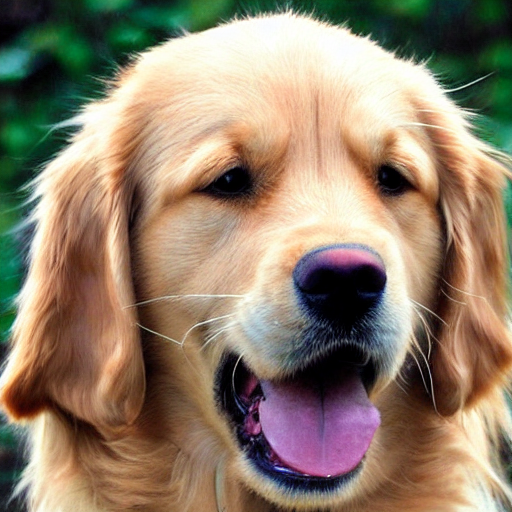} \\
{\scriptsize Labrador base} &
{\scriptsize Direct} &
{\scriptsize Adversarial} &
{\scriptsize Golden Retriever} \\
\end{tabular}
\caption{Representative successful unlearning cases of MapRoute++. In each row, the base
model produces the target concept from the target prompt, while the unlearned
model removes the target under direct and adversarial prompts. The final column
shows that a related adjacent concept remains visually recognizable.}
\label{fig:good_cases}
\end{figure}
\section{Ablations and Discussion}
\subsection{Selection of Surrogate Concepts}
\label{sec:surrogate_selection}
The fundamental difference between MapRoute and MapRoute++ lies in the selection of surrogate concepts. In MapRoute, the surrogate concepts can be arbitrary, whereas in MapRoute++, we heuristically select the surrogate concepts without revealing the target concept. These concepts are selected to redirect the diffusion model away from the target concept while preserving the model's general image-generation ability. A useful surrogate should be visually plausible, should not explicitly or implicitly contain the target concept, and should not be so close to the target that it reintroduces target-specific visual cues. At the same time, it should not be too distant or destructive, since that can damage adjacent or retained concepts. We choose surrogates using only general semantic reasoning about the target concept, without using information from the challenge dataset's indirect prompts, adversarial prompts, or adjacent-concept prompts. This avoids evaluation leakage and makes the setting conservative. This careful selection is important because poor surrogates can either fail to erase the target or cause over-erasure that harms nearby concepts, which is why we evaluate all five axes rather than direct forgetting alone.

The qualitative examples in Figures~\ref{fig:good_cases} and ~\ref{fig:failure_cases} illustrate why this careful selection matters. Successful cases show that the mapper can suppress the target concept under direct and adversarial prompts while preserving adjacent concepts. Failure cases show the two main risks of surrogate selection: the surrogate may be too weak, allowing the target concept to remain visible, or too aggressive, causing adjacent-concept degradation. These outcomes motivate evaluating unlearning with all five axes rather than only measuring direct target forgetting.

\subsection{Error Analysis}
\textbf{Successful cases.}
Figure~\ref{fig:good_cases} shows representative high-scoring concepts. These examples show that the mapper can remove the target concept while preserving nearby non-target concepts. Golf Ball and Eating have the two highest ERR scores in Table~\ref{tab:per_concept_breakdown}, and Labrador Retriever provides an object-category example with strong target forgetting and good adjacent-concept retention. The adversarial prompts are nonsensical prompt strings from the benchmark; success on this axis means that the target concept is not regenerated.

\begin{figure}[htbp]
\centering
\setlength{\tabcolsep}{2pt}
\begin{tabular}{cccc}
\includegraphics[width=0.235\linewidth]{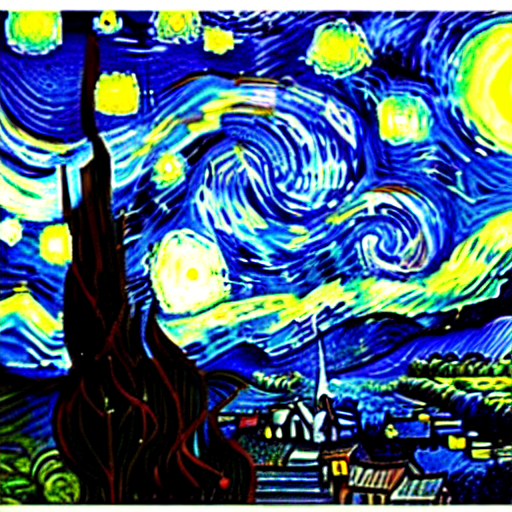} &
\includegraphics[width=0.235\linewidth]{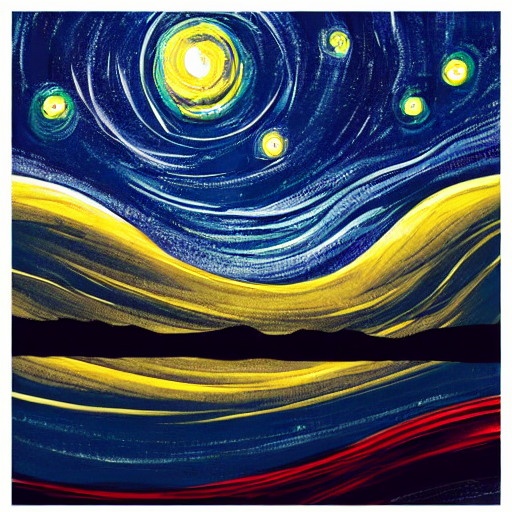} &
\includegraphics[width=0.235\linewidth]{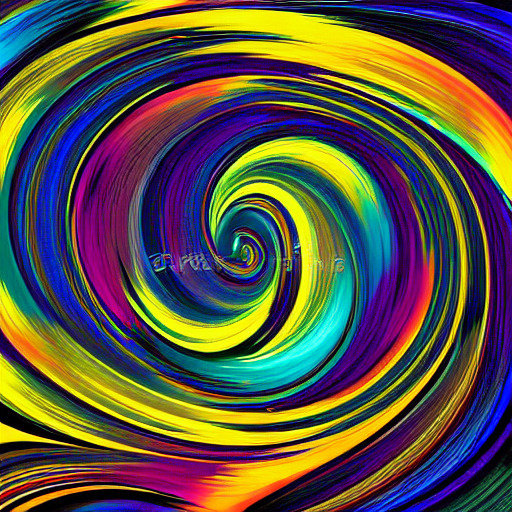} &
\includegraphics[width=0.235\linewidth]{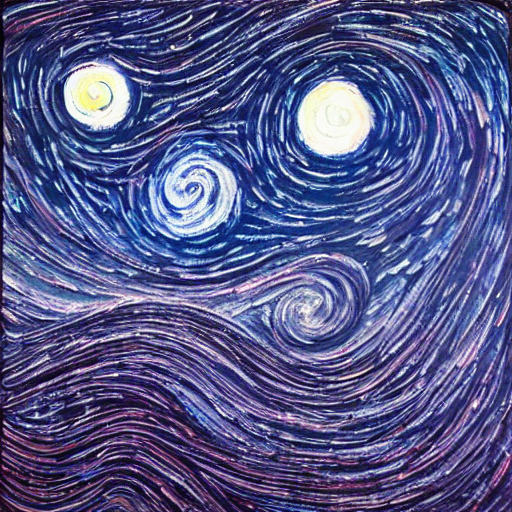} \\
{\scriptsize Van Gogh base} &
{\scriptsize Direct} &
{\scriptsize Indirect} &
{\scriptsize Adversarial} \\
\includegraphics[width=0.235\linewidth]{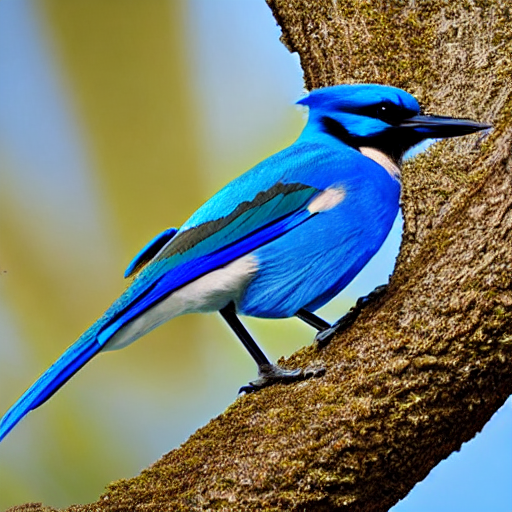} &
\includegraphics[width=0.235\linewidth]{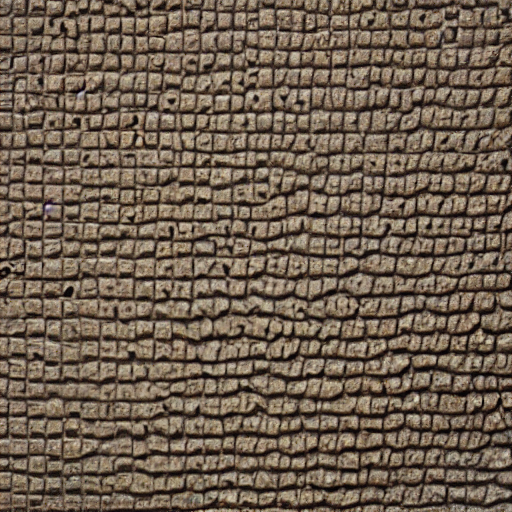} &
\includegraphics[width=0.235\linewidth]{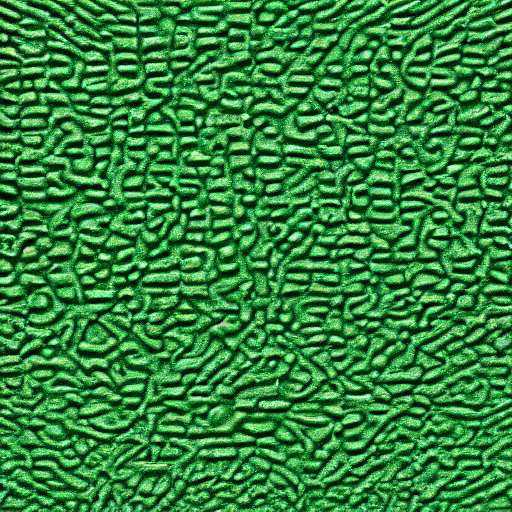} &
\includegraphics[width=0.235\linewidth]{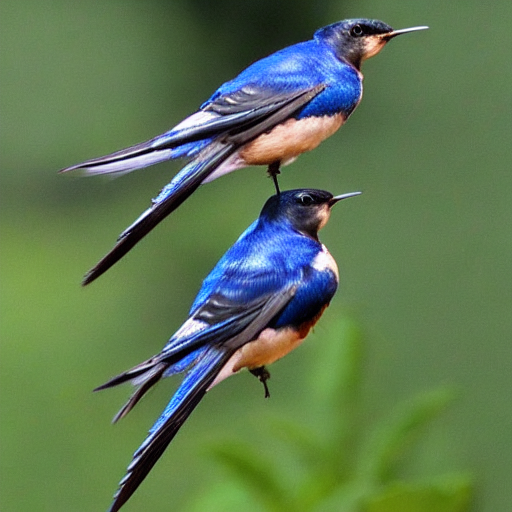} \\
{\scriptsize Florida Jay base} &
{\scriptsize Florida Jay} &
{\scriptsize Green Jay} &
{\scriptsize Adversarial} \\
\end{tabular}
\caption{Representative qualitative failures. Top row: Van Gogh remains
visually present after unlearning, including for indirect and adversarial
prompts. Bottom row: the base model produces a recognizable adjacent bird, but
the unlearned Blue Jay mapper often converts adjacent birds into texture-like
patterns; adversarial prompting can still produce bird-like imagery.}
\label{fig:failure_cases}
\end{figure}

\noindent \textbf{Failure cases.} 
Figure~\ref{fig:failure_cases} shows representative failure cases for the two most diagnostic concepts. For Van Gogh, the selected examples show that the generated images retain highly recognizable swirling brushwork and \emph{Starry Night}-like structure under direct, indirect, and adversarial prompts, illustrating failures in both forgetting and robustness. For Blue Jay, the examples show that adjacent bird prompts collapse into texture-like images, while adversarial prompts can still produce bird-like outputs, revealing both adjacent-concept damage and adversarial leakage.

\section{Conclusion}

We presented MapRoute++, a compute-efficient framework for visual concept unlearning that combines lightweight concept-specific mapper networks with a two-stage training strategy and semantic routing. By selecting relevant surrogate concepts and applying only the corresponding mapper modules at inference, MapRoute++ achieves near-zero distortion on non-target concepts while reliably suppressing designated concepts without modifying the underlying diffusion model. Experimental results on the Gen$\mu$ 2.0 benchmark demonstrate that the proposed approach consistently outperforms existing baselines in concept unlearning while preserving model utility.

\bibliographystyle{splncs04}
\bibliography{main}
\end{document}